# DOĞRU AKIM MOTORLARI İÇİN NESNELERİN İNTERNETİ TABANLI KONTROLCÜ TASARIMI

*Zeynep Özdemir[1], Mehmet Tekerek[2], Ahmet Serdar Yılmaz[1]\**

[1]Kahramanmaraş Sütçü İmam Üniversitesi, Elektrik-Elektronik Mühendisliği, Kahramanmaraş, Türkiye
[2]Kahramanmaraş Sütçü İmam Üniversitesi, Eğitim Fakültesi, Bilgisayar ve Öğretim Teknolojileri Eğitimi Bölümü, Kahramanmaraş, Türkiye

*Sorumlu Yazar: asyilmaz@ksu.edu.tr

## ÖZET

*İnternetin hayatın pek çok alanında yaygınlaşmış ve makinaların birbirleriyle haberleşmesinde imkân tanıdığı bilinmektedir. Nesnelerin İnterneti (IoT) terimi, operatörden bağımsız çalışan ve birbiriyle iletişim kuran makina ve ekipmanlardan oluşan ağlar neticesinde ortaya çıkmıştır. Sensörler, gömülü sistemler, iletişim teknolojileri ve bulutlar gibi veri depolama sistemlerinin oluşturduğu geniş alanlı ağlar haberleşebilir ve verileri paylaşabilirlerdir. Bu çalışmada, bir doğru akım motoruyla sürülen düşük hızlı mekanik bir düzen tasarlanmıştır. Aktüatörlere etki eden basıncın değişmesine bağlı olarak meydana gelen değişikliklere göre motor devrinin istenen değerde sabit tutulması amaçlanmaktadır. Yapılan çalışmada gerçek sisteme en yakın sonuçlar verecek bir modelleme ve benzetim düzeneği kurularak hız denetimi gerçekleştirilmiştir.*

**Anahtar Kelimeler:** Doğru Akım Motoru, Doğru Akım Motor Kontrolü, Nesnelerin İnterneti

# DESIGN OF INTERNET OF THINGS BASED CONTROLLER FOR DIRECT CURRENT MOTORS

## ABSTRACT

*It is known that internet have been widespread in many areas of life and enabled machines to communicate each other. The terms of Internet of Things (IoT) has emerged as a result of networks consist of machines and equipment that work independent of operator and communicate each other. Sensors, embedded systems, communication technologies and data storage systems like clouds create wide area networks that can communicate and share data. In this study, the application of the IoT on a low speed mechanical benchmark driven by a direct current motor has been designed. It is aimed to keep the motor speed fixed at desired value according to the changings occurs depending on varying the pressure acting on the actuators. In presented study, a speed control is performed by establishing a modeling and simulation mechanism that will give the closest results to the real system.*

**Keywords** DC Motor, DC Motor Control, Internet of Things,

## 1. GİRİŞ

İnternet ağları ve bilişim sistemleri, mühendislik problemlerinin çözümünde kullanılmakta ve başarılı sonuçlar elde edilmektedir. İnternet ve bilgi sistemleri başlangıçta haberleşme ve bilgi paylaşımından başlayarak günümüzde pek çok problemin çözümüne uygulanmış ve yirmi birinci yüzyılda hayatın her alanına nesnelerin interneti kavramına öncülük etmiştir ( Arslan, K. ve Kırbaş, 2016).

Nesnelerin interneti uygulamalarının mobil cihazlar ile uzaktan izlenmesi ve kontrolü, tasarlanan sistemlerin üstünlüklerini arttırmaktadır. Endüstriyel uygulamalardan pek çok alanda yaygınlaşan nesnelerin internetinde, elektrik motorlarının ve bunlarla bağlantılı mekanik elemanların kontrolü önemlidir (Gökrem L. ve Bozuklu M., 2016; Akanksha and Kathuria, 2017). Özellikle konforun öne çıktığı uygulamalarda motorlarla birlikte doğrusal aktüatörlerin birlikte kullanımı ve kurulan mekanizmanın kontrolü göz önünde bulundurulması gereken bir problem olarak görülebilir. Burada mekanizmanın ne çok hızlı nede çok yavaş çalışması beklenir. Konforu arttırmak üzere en uygun hızda çalıştırılması önemlidir. Böylece kullanım amacına uygun olarak ve kullanıcının konforunu bozmayacak şekilde çalıştırılması sağlanacaktır. Hasta bakımı, koltuk tasarımı, çalışma masası vb. alanlarda uygulanabilirliği olan bu tasarımın önemini arttırmaktadır.

Bu çalışma kapsamında; bir fırçalı doğru akım motorunun sürdüğü lineer aktuatörün yukarı aşağı doğrusal yönde hareketi ile çalışan bir düzeneğin bilgisayar destekli benzetimi gerçekleştirilmiştir. Bu benzetimde lineer aktuatörün hareket ettirdiği düzeneğin, aktuatör üzerindeki basınç etkisi izlenmiş ve tasarlanan denetleyici için giriş parametresi olarak alınmıştır. Aktuatöre uygulanan basıncın değişimi ve motor hızındaki değişime bağlı olarak aktuatörün açılma/kapanma aralığı ve açısı tespit edilmiştir. Böylece lineer aktuatörün açıklığının değişken ağırlıklarda bile aynı kalması hedeflenmiştir. Çalışmada tek bir motordan oluşan lineer aktuatör üzerinde bilgisayar benzetimi gerçekleştirilmiştir. Arduino Mega denetleyici tarafından motor sürme devresine gerekli işaretlerin uygulanması sağlanmıştır. Çalışma proteus devre analiz ve modelleme programı kullanılarak yapılmış gerçek sistem ve cihazlar ile uyumlu ve gerçek değerler kullanılarak gerçekleştirilmiştir.

## 2. GÖMÜLÜ SİSTEM TASARIMI VE ALGILAYICILAR

### 2.1. Temel Tanımlar

Son zamanlarda Arduino firmasının ürettiği çeşitli gömülü sistem kartları yanı sıra, Raspbery Pi, Orange Pi, gibi gömülü sistemler yaygın olarak kullanılmakla birlikte microchip pic, ti msp 430, ti c2000 vb gömülü sistem kartları mevcuttur. Bu çalışmada kullanım kolaylığı ve yaygınlığı bakımından öne çıkan Arduino Mega denetleyici kartı kullanılmıştır. Arduino Mega pin sayısının fazla olmasından dolayı tercih edilmiştir. Şekil 1'de çalışmada kullanılan kartın görüntüsü verilmiştir.

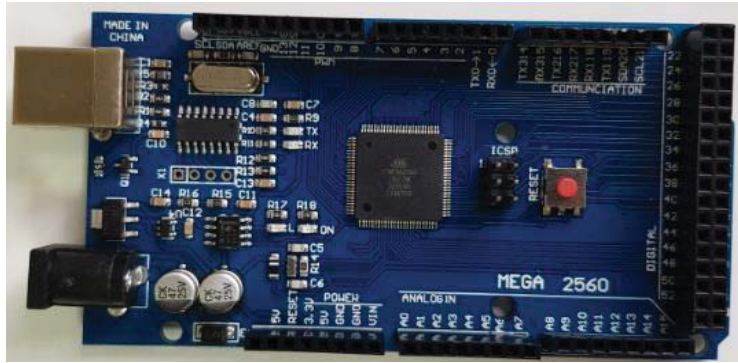

Şekil 1: Arduino Mega Mikro Denetleyici

### 2.2. Algılayıcılar

ESP8266 Modülü(Wi-Fi sensör), internete bağlanarak veri yükleme, alma ve internet üzerinden kontrol edebilme için kullanılmaktadır. ESP8266 modülü Arduino gibi geliştirici kartlarla kullanılabildiği gibi üzerinde bulunan işlemciyle programlanarak herhangi bir geliştirici karta gerek duyulmadan da kullanılabilmektedir (Kaya ve Tekin, 2018). Şekil 2.a'de bu çalışmada kullanılan ESP8266 yer almaktadır. Flex(Esnek) Sensörü, kıvrılma miktarı ile direnci değişen bir algılayıcı devre elemanıdır. Direnç, kıvrılma ile doğru orantılı olarak değişmektedir. Tek-yönlü veya çift-yönlü olarak oluşturulabilmektedir. Kıvrılma açısı arttıkça esneklik sensörünün direnci de artmaktadır (web1,2019). Bu çalışmada Şekil 2.b de görülen, flex sensörü basınç algılaması ve gördüğü basınca bağlı olarak atadığı direnç değeriyle sistemin kontrolünü sağlayabilmek için kullanılmıştır.

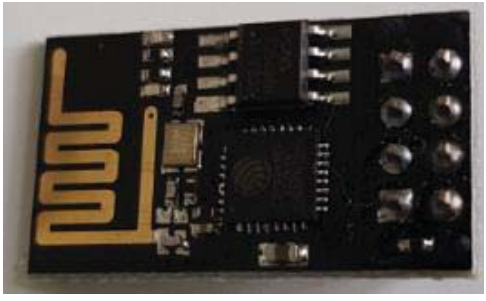 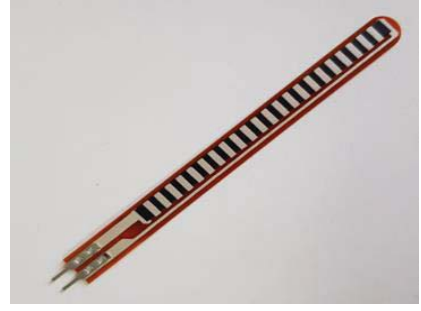

(a) (b)

Şekil 2.a: ESP8266 Wifi Algılayıcı, b: Flex (Esnek) Basınç Sensörü

## 2.3. Doğru Akım Motoru ve Sürücü

Doğru akım motorları, endüstriyel uygulamalardan günlük hayattaki uygulamalara kadar oldukça yaygın bir kullanım alanına sahip elemandır. Kontrol edilebilmesinin kolaylığı ve küçük boyutlu uygulama noktalarında rahatça kullanılabilirliği üstünlük olarak karşımıza çıkmaktadır. Doğru akım motorları, pozisyon kontrolü (Dogman ve Boz, 2002) ve hız kontrolü (Nasreldin vd, 2017) için diğer motor türlerine göre kullanışlılığı ve basitliğiyle öne çıkmaktadır.

Bu çalışmada kullanılan ve Şekil 3.a'da verilen, redüktörlü motor, 131.25: 1 metal şanzımanlı güçlü bir 12V fırçalı DC motordur. Şanzıman çıkış milinin devir başına 8400 sayımına karşılık gelen motor şaftı devri başına 64 sayım çözünürlüğü sağlayan entegre bir dörtlü kodlayıcıdır. Bu birimler 16 mm uzunluğunda, 6 mm çapında D şeklinde bir çıkış miline sahiptir. Bir manyetik diskin motor şaftının arka çıkıntısındaki dönüşünü algılamak için iki kanallı bir Hall efekti kodlayıcısı kullanılmıştır. Dörtlü kodlayıcı, her iki kanalın iki kenarını da sayarken motor şaftının devri başına 64 sayımlık bir çözünürlük sağlamaktadır. Mikrodenetleyicilerin voltaj çıkışları DC motorları veya step motorları direkt olarak kontrol etmek için motor sürücü devreler kullanılmıştır. Bu çalışmada şekil 3.b'de görülen L298 motor sürücü devresi simülasyon ortamında kullanılmıştır.

Elektrikli doğrusal aktuatör, düşük voltajlı bir DC motorun dönme hareketini, doğrusal harekete dönüştüren bir cihazdır (web2,2019). Şekil 3.c'de tipik bir doğrusal aktuatör görülmektedir.

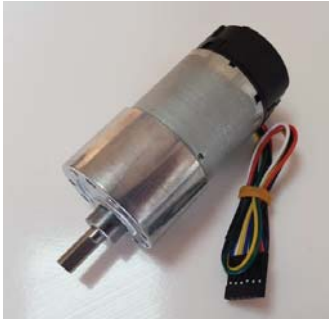 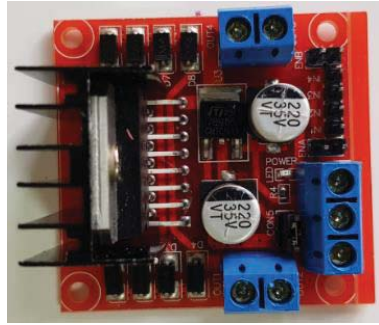 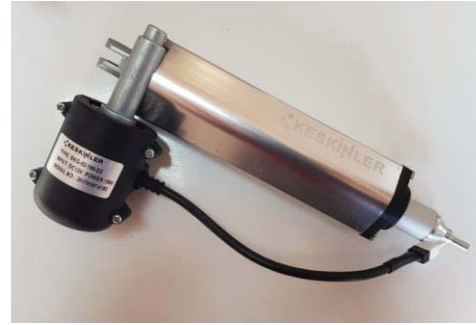

(a) (b) (c)

*Şekil 3.a:Redüktörlü Motor, b: L298 Motor Sürücü, c: Lineer Aktuatör*

## 3. BİLGİSAYAR DESTEKLİ MODELLEME ve BENZETİMLER

Labcenter Electronics firmasının bir ürünü olan Proteus Desing Suite programı ISIS ve ARES olmak üzere iki alt programdan oluşmaktadır. ISIS'te elektronik alanında devre çizimi, simülasyon (modelleme), animasyon (canlandırma) gerçekleştirilebilmektedir. ISIS'te hazırlanmış olan devreler ARES ortamına aktarılarak manuel veya otomatik baskılı devre çizimi (PCB) ve 3D görsel modelleme yapılabilmektedir.

*Doğru Akım Motorları için Nesnelerin İnterneti Tabanlı Kontrolcü Tasarımı*

*Zeynep Özdemir, Mehmet Tekerek ve Ahmet Serdar Yılmaz*

Bu programla elektrik–elektronik devre şemaları çizilip, bu çizimler yazılı dokümanlarda kullanılabilmektedir. Program; kullanıcıları laboratuvardaki kablo karmaşasından kurtarır ve aynı zamanda, oluşturulan devreyi gerçek ölçütlerle simüle ederek, gerçek devreye çok yakın sonuçlar alınmasına yardımcı olabilmektedir (web2,2019 ; web3,2019).

Bu çalışmayı ISIS ekranı üzerinde tasarlayabilmek için öncelikle arduino kütüphanesini indirerek Proteus kütüphanesine eklememiz gerekmektedir. Bu çalışmada, nesnelerin interneti tekniği ile enkoderli bir DC motorun kontrolünün basınca duyarlı olarak sağlanması hedeflenmiştir. Bağlantı sonrasında esnek sensörden analog olarak okunan 1 veya 0 göre motor aktif veya pasif olacaktır. Aktif olması durumunda motor yol almaya başlar. Simülasyon devresi Şekil 4'te görülmektedir.

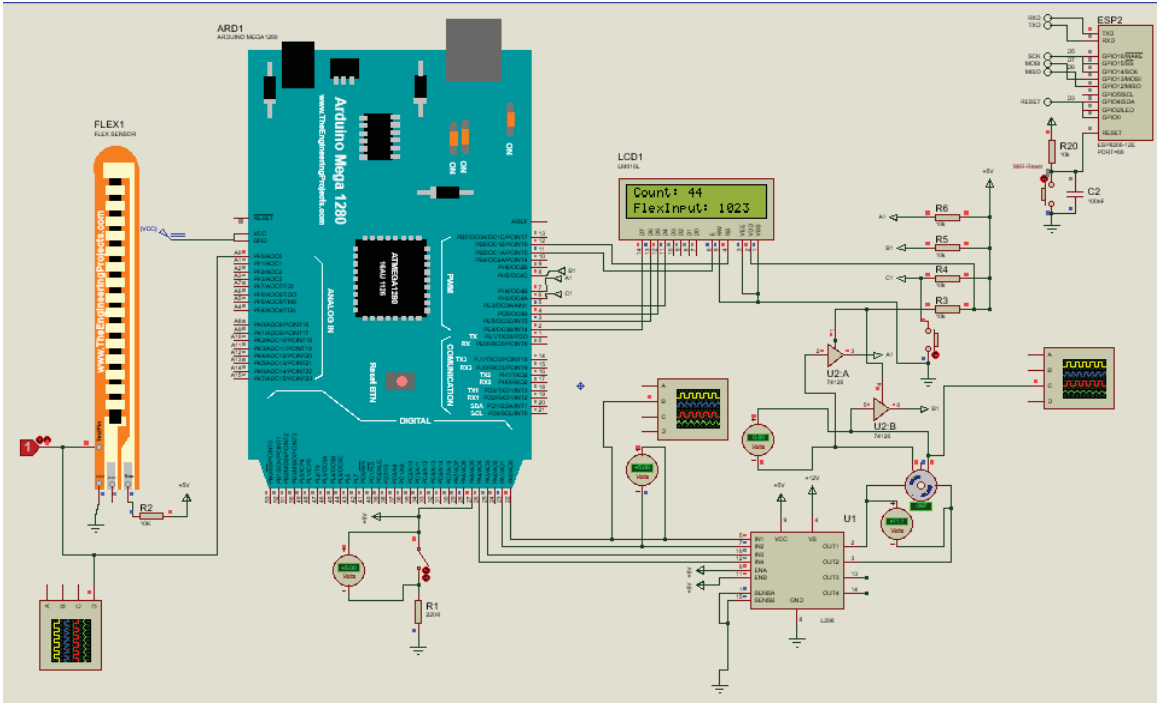

*Sekil 4: Simülasyon devresi*

Motor enkoderinin her bir turu sayılarak LCD Ekrana yazılmaktadır. Böylece motor enkoderi üzerindeki göstergede yer alan RPM değerine ne kadar süre sonra geldiği simülatördeki bilgilerden elde edilmiştir. Motor +12V besleme gerilimine ihtiyaç duyduğu için doğrudan Arduino Megaya bağlamak yerine motor sürücüsü kullanılmıştır. Motor enkoderli olduğu için kendi beslemesi yanında enkoder beslemesi ve 1 adet enkoder sinyal çıkışı vardır. Proteusta enkoderli DC motor bağlantısı, DC motora bağlı Rotary Encoder şeklinde yapılır.

Şekil 4'te görüleceği gibi ESP8266 ISIS ekranına eklenerek TXD ve RXD pinleri, Arduino Meganın TXD ve RXD pinlerine bağlantısı yapılmıştır. Proteus web tarayıcı arayüzü üzerinden bir push button ile esnek sensörün kontrolü gerçekleştirilmiştir. Böylece motor sürücü devresi aktif edilmiştir ve enkoderli DC motor yol almaya başlamıştır. Motor sürücüsü olarak kullanılan L298'in IN1ve IN2 pini Arduino Meganın digital 22. ve 23. pinlerine bağlanmıştır. Sürücünün ENA ve ENB pinlerine +5V bağlanılması halinde OUT1- OUT2 ve OUT3- OUT4 pinleri aktiflececektir. GND bağlanılması halinde pasif olacaklardır. Burada ENA ve ENB pinlerine +5V bağlantısı yapılmıştır. Fakat OUT3-OUT4 çıkışları kapatılmıştır. OUT1-OUT2 pinleri enkoderli DC motorun motor besleme girişlerine bağlanmıştır. Böylece IN1 ve IN2 pinlerinden gelen veriye göre motor yönü kontrol edilebilmektedir. Şekil 4'den görüleceği gibi ESP8266 ISIS ekranına eklenerek TXD ve RXD pinleri, Arduino Meganın TXD ve RXD pinlerine bağlantısı yapılmıştır. Proteus web tarayıcı arayüzü üzerinden bir push button ile esnek sensörün kontrolü gerçekleştirilmiştir. Böylece motor sürücü devresi aktif edilmiştir ve enkoderli DC motor yol almaya başlamıştır. Şekil 5'de her iki sensör birlikte görülmektedir.

*Doğru Akım Motorları için Nesnelerin İnterneti Tabanlı Kontrolcü Tasarımı*

*Zeynep Özdemir, Mehmet Tekerek ve Ahmet Serdar Yılmaz*

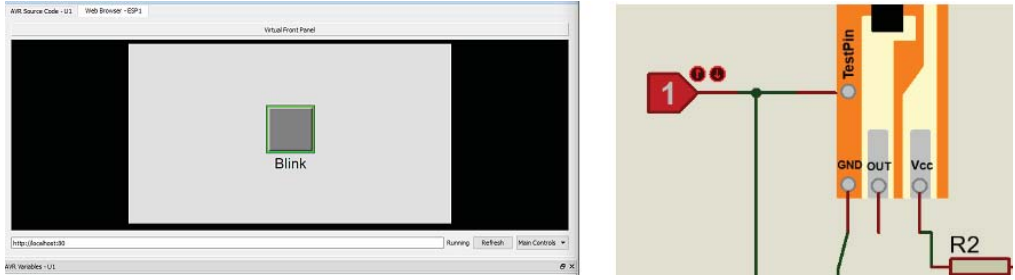

*Şekil 5: Kullanımdaki ESP8266 ve Esnek Sensör*

Simülasyon sonuçları yine program içerisinde bulunan osiloskop eklentisi kullanılarak elde edilmiştir[18]. Bu çalışmada devre elemanlarında gerilimdeki zamana bağlı değişimi görebilmek için osiloskop bağlantısı kullanılması gerekmektedir. Voltmetrede yalnızca gerilim büyüklüğü bilgisi elde edilmektedir. L298 girişine, enkoderli DC motor girişine ve flex sensör girişlerine osiloskop bağlanmıştır.

Şekil 6.a'da L298 motor sürücüsüne gelen +5V giriş sinyalinin osiloskop ekran görüntüsü verilmiştir. Motor enkoderi A ve B çıkışları, 0-Vcc seviyesinde, aralarında yaklaşık olarak 90° faz farklı bulunan kare dalga şeklinde çıkış verir. Bu dalgaların frekansı motorun hızını, sırası ise motorun dönüş yönünü vermektedir.

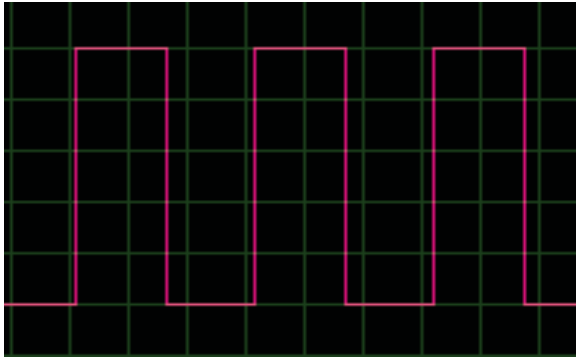
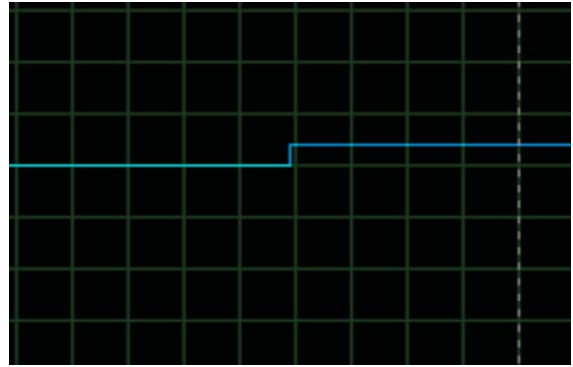

(a)             (b)

*Şekil 6.a: Enkoder A çıkışı Voltaj Grafiği, b: Motor Sürücü Voltaj Grafiği*

Esp8266 üzerinden flex sensör aktif yapıldığında sahip olduğu +5V şekil 7.a'da görülmüştür. Esp8266 ile esnek sensör bağlantısı kurulmadığında yani esnek sensör pasif iken 0V'a sahip olduğu şekil 7.b'de görülmektedir.

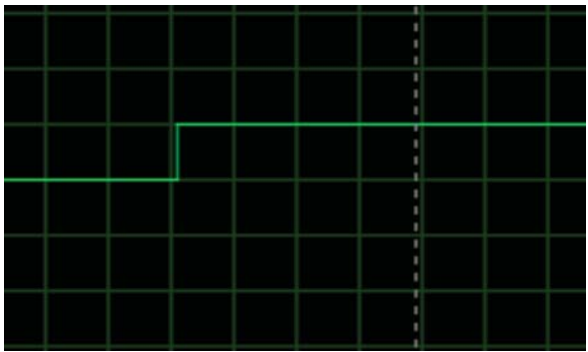
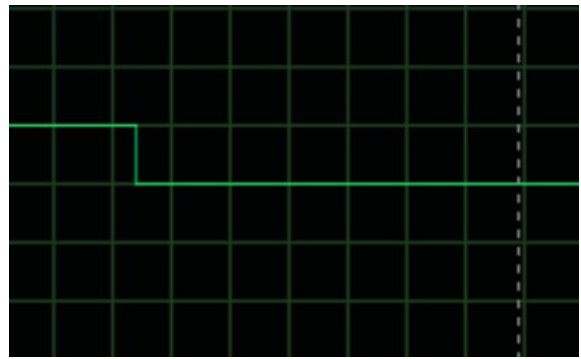

(a)             (b)

*Şekil 7.a: Flex Sensör Basınç Uygulandığında Voltaj Grafiği, b: Flex Sensör Basınç Uygulanmadığında Voltaj Grafiği*

Bu durumda L298 üzerine bağlanan voltmetre üzerinden motor sürücünün de pasif olduğu enkoderli DC motorun ise 0 voltajda olup yol almadığı görülmüştür. Şekilden de anlaşılacağı üzere öncelikle motor sürücüsü sonrasında da motorun +5V voltaj gerilimine sahip olduğu ve yol aldığı sırasıyla görülmüştür.

## 4. SONUÇ VE ÖNERİLER

Bu çalışma herhangi bir koşul üzerinden DC motor kontrolünün IoT aracılığıyla gerçekleşebileceği ve pozisyon bilgisinin elde edilebileceği temeli üzerinedir. Çalışmada hedeflenen wi-fi bağlantısı ESP8266 ile arduino da yazılan kod üzerinden sağlanmıştır. Basınç kontrolüyle DC motora PWM sinyali gittiği osiloskop grafiklerinden gözlemlenmiştir. Bağlantının doğru olduğuna kanaat getirerek DC motorun pozisyon bilgisi, yönü ve RPM'i motorun üzerindeki sayaçta görülmüştür. Böylece nesnelerin interneti tekniği ile söz konusu DC motorun kontrolü sağlanmıştır. Çalışma kapsamında gerçekleştirilen benzetim uygulaması gelecekte yapılacak deneysel çalışmalara ışık tutacaktır.

DC motor üzerinden gerçekleştirilen bu benzetim sayesinde, uygulamanın gerçek bir uygulamaya dökülebilir olduğunu görülmüştür birçok farklı sensör, röle ve anahtarlama devreleriyle de gerçekleşebileceği anlaşılmıştır. Enkoderli DC motor ile Lineer Aktüatörü birleştirip akuple bir sistem oluşturarak IoT ile kontrol edilebilen basınca duyarlı bir lineer aktüatör oluşumu sağlanabilecektir. Bunun bir ileri seviyesi olarak aktüatör sayısını arttırarak daha çok koşul üzerinden aktüatörlerin kontrol ve pozisyon bilgisini alabilmek mümkün olacaktır. Daha sonrası ise belirli koşullar yine göz önünde bulundurularak, aktüaörlerin birbirleriyle iletişime geçerek çeşitli kombinasyonlarda çalışmasını sağlamak olacaktır. Ayrıca gömülü sistem olarak arduino temelli kartlar dışında özellikli kartların kullanılması da mümkündür.

Bu çalışma simülasyon tabanlı bir çalışma olup, motor sayısı arttırılarak hareketli masa, yatak, sandalye gibi ürünlere uygulanabilir niteliktedir. Çalışmanın devamında deneysel olarak uygulanması hedeflenmektedir. Arduino çok yaygın olarak bilinen temel bir mikrodenetleyici olup kapasitesi nispeten sınırlıdır. Bundan sonraki çalışmalarda Raspberry Pi, Orange Pi, PIC vb. mikrodenetleyiciler kullanılarak çok sayıda motorun kontrolü sağlanacaktır.

## 5. KAYNAKLAR